\pdfoutput=1
\documentclass[11pt]{article}

\usepackage{acl}

\usepackage{times}
\usepackage{latexsym}

\usepackage[T1]{fontenc}
\usepackage[utf8]{inputenc}

\usepackage{microtype}

\usepackage{inconsolata}

\usepackage{times}
\usepackage{latexsym}
\usepackage{multirow}

\usepackage[T1]{fontenc}
\usepackage{tikz}
\usepackage{tabularx}
\usepackage{amsmath}
\usepackage{amsthm}
\usepackage{amssymb}
\usepackage{paralist}
\usepackage{booktabs}
\usepackage{todonotes}
\usepackage[capitalise,noabbrev]{cleveref}

\usepackage{xspace}

\usepackage{makecell}

\title{LUCID: LLM-Generated Utterances for Complex and Interesting Dialogues}

\author{\makecell{Joe Stacey$^{1}$\thanks{Work undertaken while author was an intern at Apple} ~~~~~~~ Jianpeng Cheng$^{2}$ ~~~~~~ John Torr$^{2}$ ~~~~~~ Tristan Guigue$^{2}$
\\
Joris Driesen$^{2}$ ~~~~~~ Alexandru Coca$^{3*}$ ~~~~~~ Mark Gaynor$^{2}$ ~~~~~~ Anders Johannsen$^{2}$} \\
$^{1}$Imperial College London \hspace{4mm}
$^{2}$Apple ~~~~~   \hspace{4mm}
$^{3}$University of Cambridge \\
\texttt{\makecell{j.stacey20@imperial.ac.uk, ac2123@cam.ac.uk\\ \{jianpeng.cheng,jtorr,tguigue\}@apple.com\\
\{joris\_driesen,mgaynor,ajohannsen\}@apple.com}}}

\begin{document}
\maketitle

\begin{abstract}
Spurred by recent advances in Large Language Models (LLMs), virtual assistants are poised to take a leap forward in terms of their dialogue capabilities. Yet a major bottleneck to achieving genuinely transformative task-oriented dialogue capabilities remains the scarcity of high quality data. Existing datasets, while impressive in scale, have limited domain coverage and contain few genuinely challenging conversational phenomena; those which are present are typically unlabelled, making it difficult to assess the strengths and weaknesses of models without time-consuming and costly human evaluation. Moreover, creating high quality dialogue data has until now required considerable human input, limiting both the scale of these datasets and the ability to rapidly bootstrap data for a new target domain. We aim to overcome these issues with LUCID, a modularised and highly automated LLM-driven data generation system that produces realistic, diverse and challenging dialogues. We use LUCID to generate a seed dataset of 4,277 conversations across 100 intents to demonstrate its capabilities, with a human review finding consistently high quality labels in the generated data\footnote{Please see http://github.com/apple/ml-lucid-datagen for the data and project code}. %The generated conversations include a wide range of challenging phenomena and diverse user behaviour, identifiable via a set of turn-level tags. Finally, we provide separate test sets for seen and unseen intents, allowing for convenient out-of-distribution evaluation. We release both the data generation code  and the dataset itself.
\end{abstract}

\section{Introduction}
\begin{figure}
\begin{center}
    
\includegraphics[width=\columnwidth]{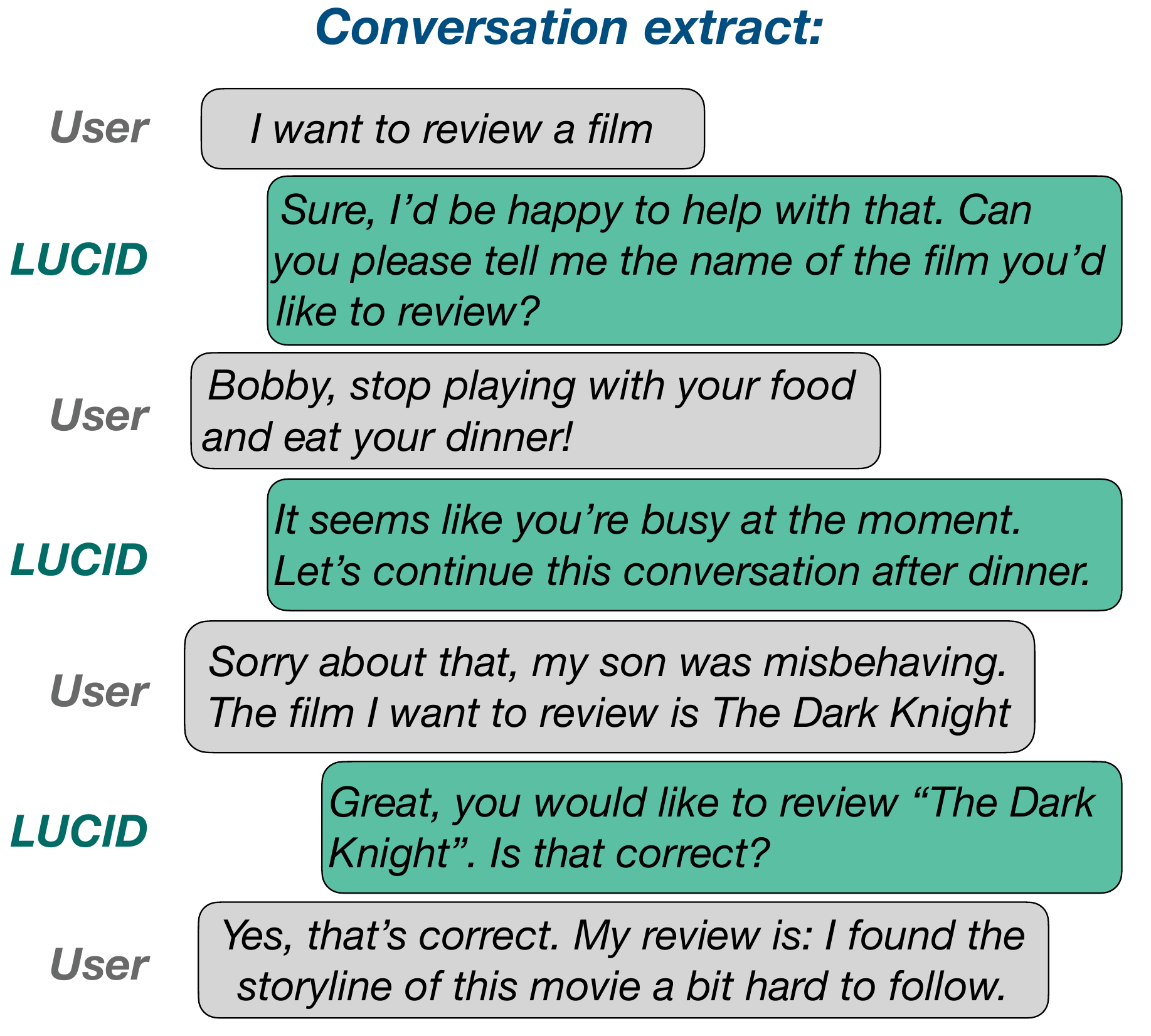} 
    \caption{An extract of a LUCID conversation containing a challenging phenomenon. In this case, the second user response is most likely to be from an overheard conversation rather than providing the desired slot value.} \label{single_extract}

\end{center}
\end{figure}

As AI virtual assistants become more sophisticated, there is an increasing need for dialogue datasets with more challenging conversational phenomena for both fine-tuning and evaluation. Existing datasets include multi-turn, multi-intent and multi-domain conversations \cite{SGD, MultiWOZ}, in addition to multi-lingual datasets \cite{PRESTO, MASSIVE_by_Jack_from_Alexa, MultiWOZ_Multilingual, MTOP_dataset}. However, in each case, the number of intents covered is relatively small. Moreover, the conversational phenomena included in these datasets are often limited in scope\footnote{The PRESTO dataset \citep{PRESTO} does explicitly label specific conversational phenomena, but to the best of our knowledge it is unique in this respect}. Additionally, current machine-to-machine data collection methods still involve varying degrees of human involvement, with humans paraphrasing machine generated templates into natural language, and/or manually crafting plausible sequences of intents as dialogue outlines \citep{M2M_paper, SGD}. %Even this reduced level of human involvement in the data generation process, relative to even more onerous human-to-human or human-to-machine approaches, is costly and time consuming, and difficult to scale to new target domains. 

To overcome these issues, we introduce LUCID, \textbf{L}LM-generated \textbf{U}tterances for \textbf{C}omplex and \textbf{I}nteresting \textbf{D}ialogues. LUCID is composed of a pipeline of modularised LLM calls that create realistic, accurate and complex data, allowing the data generation process to scale to more intents, slots and challenging conversational phenomena. LUCID involves automated intent generation, with a mock back-end\footnote{The mock back-end converts intents into Python classes, which are then instantiated as objects} created for each intent. This mock back-end then interacts with LLM-based user and system agents, generating dialogues without the need for human annotation. %, avoiding the need for human annotation.

%To the best of our knowledge, this is the first work generating task-oriented dialogue data that does not rely on human annotations.  %Adding new conversational phenomena to the system is a simple matter of adding a set of imperatives written in natural language to a python file, for example "When Lucid asks about \{arg1\}, don't answer, but instead say something to another person who is in the same room" (see \cref{single_extract}).

% This includes a range of conversational phenomena that require a model to distinguish irrelevant user utterances, for example for sarcastic responses or when other conversations are overheard by the virtual assistant.
% Introduce LUCID acronym

Ensuring data quality is a central challenge for a machine-to-machine generation process. We address this issue by breaking down the generation process into a pipeline of multiple, simpler LLM calls, thereby compartmentalising the data generation task into manageable steps that an LLM can consistently perform accurately. In addition, we use multiple LLM-based validators which discard conversations that \textit{might} contain an issue. Our \textit{if in doubt, discard} philosophy ensures a high quality standard for the data being created. %Finally, we use sampling throughout our generation process to ensure variety in the length and content of each conversation. To showcase the quality, realism and sophistication of the data generated by LUCID, we release a seed dataset of 92,699 turns.

We release the data generation code to enable large scale data generation across different intents and domains, with the option of adding additional, complex conversational flows. We also provide training data, validation data, and two tests sets, a test set for seen intents, and an additional test set for unseen intents, allowing for convenient out-of-distribution evaluation.

%To summarise our contributions: 1) We introduce a highly automated methodology for producing high quality LLM-generated conversations. 2) We provide a challenging dataset of 4,277 dialogues across 100 intents, including more labelled conversational phenomena than previous dialogue datasets. 3) We provide two tests sets, a test set for seen intents, and an additional test set for unseen intents, allowing for convenient out-of-distribution evaluation. 4) We release the data generation code to enable large scale data generation across different intents and domains, with the option of adding additional, complex conversational flows.

\section{Related Work}
\subsection{Task Oriented Dialogue Datasets}
The most popular approach for creating dialogue datasets involves human-to-human interactions, with user annotators interacting with Wizard of Oz (WoZ) annotators \cite{MultiWOZ, frames_dataset, zhu-etal-2020-crosswoz, WoZ_in_car_navigation, woz2.0}. While using human annotators can create diverse, large scale datasets, this is done at a considerable cost, with expert annotators required for accurate dialogue annotations. User annotators follow a generated conversation plan \citep{MultiWOZ, frames_dataset, zhu-etal-2020-crosswoz}, guiding their interactions with the WoZ agent. We find that even in a purely machine-to-machine setup, generating conversation plans for each dialogue remains an effective way to ensure conversational variety. %LUCID extends these plans to also specify how and when any complex conversational phenomena should be included in the dialogue. %Human annotation is also used in multi-lingual datasets, translating a text into another language \cite{MASSIVE_by_Jack_from_Alexa}.
\subsection{Automated Data Collection Methods}
To reduce the workload of annotators, dataset collection is becoming increasingly automated. A popular approach is to generate conversation outlines, which are then converted into natural language by annotators \cite{M2M_paper, SGD, M2M_multilingual} or using natural language templates \citep{bordes2017learning}. As these conversation outlines are simulated based on hard-coded rules, this can limit the diversity of the user behaviour.

%Natural language templates can be used to convert these simulated outlines into natural language \citep{bordes2017learning}, although at the expensive of fluent and varied conversations. 

Human involvement in automated data generation includes ensuring the quality of the dataset, paraphrasing user and agent responses \citep{M2M_paper, SGD}, providing semantic annotations \citep{PRESTO, MultiWOZ}, outlining the sequences of user intents \citep{SGD}, and identifying out of scope or incoherent examples \citep{PRESTO}. We show that, with recent advances in language modelling \cite{openai2023gpt4, InstructGPT}, by reducing the data generation task into manageable steps, and using our \textit{if in doubt, discard} validation methodology, it is now possible to achieve the same quality in an almost entirely machine-to-machine generation process. Parallel work by \citet{liu2024toad} also introduces an automated method for generating task-oriented dialogue data. While we focus on the accurate labelling of challenging and diverse conversations, \citet{liu2024toad} consider a variety of user personas with different styles of system responses.

%\citet{SGD} introduce a schema-guided paradigm, which we adopt for LUCID. In this approach, predictions for an intent are made over a defined set of slots and constraints defined in the intent's schema. This approach allows us to measure out-of-distribution performance on intents that were unseen during training, providing a model with only the schema for these intents. We provide both an in-distribution and out-of-distribution test set. Similar to our approach of 
% Also talk about how PRESTO produces labels... mostly for correction, and these are human labelled.

See \cref{sec:related_work_llm_gen_data} for related work generating data with LLMs for other tasks.

\section{Method}
\begin{figure}
\begin{center}
    
\includegraphics[width=\columnwidth]{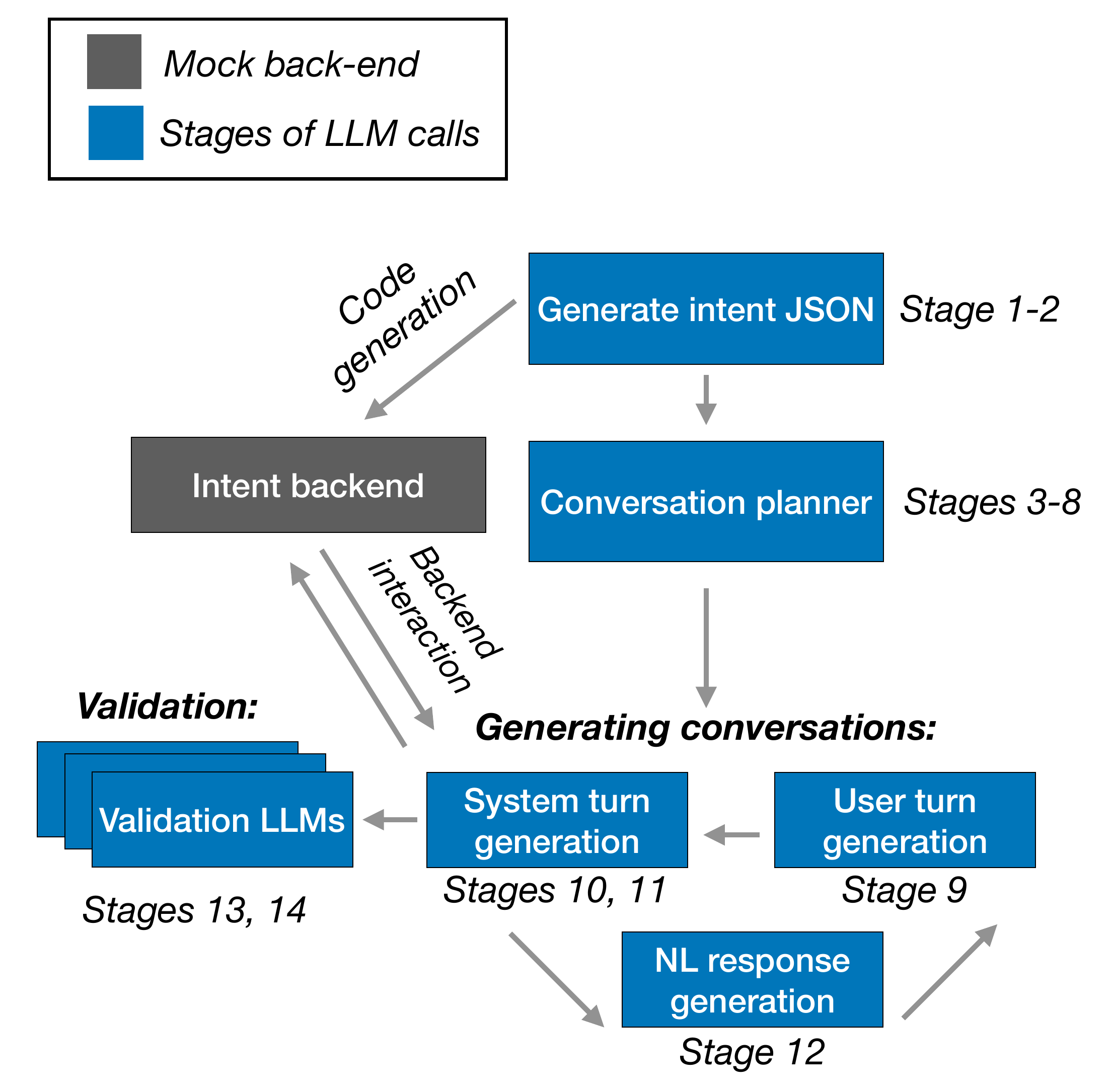} 
    \caption{The stages in the LUCID data generation, generating intents (\textit{stages 1-2}), planning conversations (\textit{stages 3-8}), generating the conversations (\textit{stages 9-12}) and validating the system predictions (\textit{stages 13-14}).} \label{method_diagram}

\end{center}
\end{figure}

LUCID decomposes the data generation process into \textbf{14} individual LLM calls, described here as \textit{stages}, creating manageable steps that LLMs can perform accurately. Alongside our \textit{if in doubt, discard} validation, reducing the complexity of each LLM call helps to ensure the quality of our generated data.
The data generation process consists of four main components (see \cref{method_diagram}): the generation of intents (\textit{stages 1-2}), a conversational planner (\textit{stages 3-8}), turn-by-turn generation of conversations (\textit{stages 9-12}), and our validation process (\textit{stages 13 and 14}). The turn-by-turn data generation involves a User LLM agent interacting with a System LLM agent, which in turn interacts with a mock back-end created for each intent.

%A further challenge when using an LLM is ensuring a variety of conversations. Even when reducing the temperature value, conversations can be repetitive, especially over such a large dataset. To avoid this, we introduce a variety of sampling procedures throughout the data generation process.

\subsection{Intent Generation}

%- Need to get across how this means we can include a larger number of intents (and domains) than previous datasets. E.g. no looking at plausible paths, as in SGD.
Schema for each intent are generated by an LLM, using a short human-authored natural language description of the intent (\textit{stage 1})\footnote{Our code also allows intent schema to be created manually}. 
%While we initially experimenting with generating intent descriptions with an LLM, as writing intent descriptions is not a time intensive task we reverted to humans providing these descriptions. 
Using these descriptions, LUCID calls an LLM to generate the intent and slot names, as well as the data type of each slot and whether it is mandatory or optional. In total, 100 intents are generated across 13 domains (see \cref{sec:intent_generation} for a detailed breakdown of how transactional and query intents are generated). %A manual review of the generated intents for LUCID led to one transactional intent being discarded (see \cref{sec:intent_generation}). 

The next stage (\textit{stage 2}), involves generating plausible values for each slot. We use these slot values as a starting point for our conversation planner, helping to encourage varied conversations.

\subsection{Conversation Planner}
The conversation planner provides instructions that guide a user LLM agent down certain types of conversational flows. The planner specifies: 1) the sequence of intents, 2) the slot values for each intent, and 3) any complex conversational phenomena that must be included, specifying when and how these phenomena should be incorporated. This creates a plan that the user must adhere to, reducing the complexity of the data generation task, while also ensuring variety in the generated conversations. This plan is communicated to the User LLM at each turn through a series of conversation rules. %, specifying the next intent (and its corresponding slot values) that should be requested, in addition to specifying the next complex conversational phenomena. The planner creates the conversation plan at the start of each conversation, involving a pipeline of different stages (\textit{stages 3-8}).

The planner also decides the sequence of intents that will be included in a conversation  (\textit{stage 3}). Depending on the primary intent used to start the conversation, the planner then decides which intents are likely to follow this intent, with the aim of creating both varied and realistic conversations. See \cref{sec:planner_sampling} for further details about the planner. %To ensure variety in the generated conversations, the planner makes further use of sampling, choosing how many intents should be provided in the conversation, which optional slots should be discussed, the conversational phenomena (both happy and unhappy paths) that should be included, and which slots and intents any conversational phenomena should be applied to.

\subsubsection{Generating Slot Values}
The slot values chosen by the planner substantially impact the conversations, and as a result, we have multiple, separate stages for generating the slot values (\textit{stages 4, 5, 6 \& 7}). This process involves updating the slot values to make sure these are realistic and coherent (\textit{stage 4)}, generating a reason why the user wants to perform any subsequent intents (\textit{stage 5}), and generating slot values for the subsequent intents based on this justification (\textit{stage 6}). Finally, an LLM updates the slot values across every intent in the conversation to ensure they are consistent and realistic when considered collectively (\textit{stage 7}).

We additionally use an LLM to generate realistic entities to be returned after any queries (\textit{stage 8}).

\subsection{Generating Conversations}
Conversations are generated turn-by-turn with a User LLM interacting with a System LLM, which in turn interacts (via Pythonic function calls and variable assignments) with a mock back-end for each intent. A Response LLM then communicates natural language responses back to the user. The user behaviour is governed by the conversational rules created by our planner, shaping the outcome of the conversations.%The conversation plan provides instructions to the User LLM about how to behave, providing details of the next intent to be requested, along with the desired slot values for that intent. These instructions are contained in conversation rules that govern the user behaviour.

Conversations start with an utterance from the User LLM (\textit{stage 9}), which is then interpreted and labelled by the System LLM (\textit{stage 10}). Initially, no string slot values are predicted. These values are predicted in an additional stage (\textit{stage 11}), where an LLM is instructed, where possible, to choose the string values from spans of the user utterance (avoiding hallucinations). %Predicting the string slot values is a challenging task, with accurate predictions being essential for high quality labelling. 
The predicted semantic labels then interact with a mock back-end for each intent. The mock back-end then informs the System LLM about any missing slots or whether confirmation is required for the intent. Finally, the Response LLM responds back to the user (\textit{stage 12}), requesting any additional slots or asking the user for confirmation.
%Once the user finishes requesting the intents outlined in the conversation plan, they signal they want to end the conversation.

\subsection{LLM-based Validation}
We implement an LLM-based validation process to ensure reliable and consistent labelling in the conversations, using our principle of \textit{if in doubt, discard}. First, based on the observation that the system LLM is more uncertain about incorrect predictions, we repeat the system predictions twice (using a temperature value of 0.7), and abort the conversation if the three predictions are not identical (\textit{stage 13}). Additional validation is then performed by another LLM (\textit{stage 14}) which also labels the user requests, except this time with access to the conversation rules that the user is following. These predictions must also exactly match the original System LLM predictions, otherwise the conversation is aborted. The validation in stages 13 and 14 is performed before the string slot values are predicted, avoiding conversations being aborted when these slot values have slightly different phrasing.
%While this additional validation step can hallucinate slot values not yet mentioned by the user, there are other occasions when it better interprets what the user is describing. <-- Move to analysis section when I have something more concrete. 
% These validation steps are performed before string slot values are predicted by the model. 
Further validation is described in \cref{sec:validation_info}.
%Further validation is provided for our conversational phenomena, described in \cref{sec:conv_phenomena}, in addition to a range of post-processing rules introduced after a qualitative analysis (described in \cref{sec:post_processing}). In total, 56\% of the conversations generated passed our validation checks (see \cref{sec:validation_breakdown} for further information). 

%To avoid wasting valuable conversational data, we salvage the prefix of an aborted conversation up to the point where the validation error was identified\footnote{To avoid overly short conversations, we do this only if at least one intent has been performed already or at least 10 turns have occurred}. In these cases, we truncate the conversation, sampling LLM generated natural language responses that justify interrupting the conversation.

\subsection{Introducing Additional Conversational Phenomena} \label{sec:conv_phenomena}

To make interesting, diverse and challenging conversations, we introduce a wide range of conversational phenomena which are labelled automatically at a turn level (see \cref{unhappy_paths}). These phenomena include \textit{sarcastic} or \textit{irrelevant} replies, or cases where the user is \textit{overheard} in another conversation.  LUCID also contains examples where a user corrects themselves, either within a turn (\textit{in-turn correction}) or in a later turn (\textit{correction}). Alternatively, a user may cancel an intent (\textit{cancellation}) or delay confirming the intent until a future turn (\textit{delay confirmation}). Our conversational phenomena also include cases where the virtual assistant requests a value for one slot, but the user responds about a different slot (\textit{respond different slot}). Finally we also include ASR early end errors (\textit{ASR-early end}), where the LLM produces truncated slot values where the user text is abruptly cut off. See \cref{unhappy_path_numbers} for the full distribution of these phenomena.%The user can also reply about multiple slots when asked about the value of a single slot (\textit{respond\_multiple\_slots}). 

%The conversation rules are sampled by the planner for each conversation, before the planner assigns these phenomena to specific intents (and slots) within the conversation. In addition to the \textit{unhappy path} conversational phenomena discussed, we also introduce different \textit{happy paths} via conversation rules to introduce variety within conversations. These rules state whether or not the user should provide some, all or no slot values when expressing the intent for the first time.

\begin{figure}
\begin{center}

\includegraphics[width=\columnwidth]{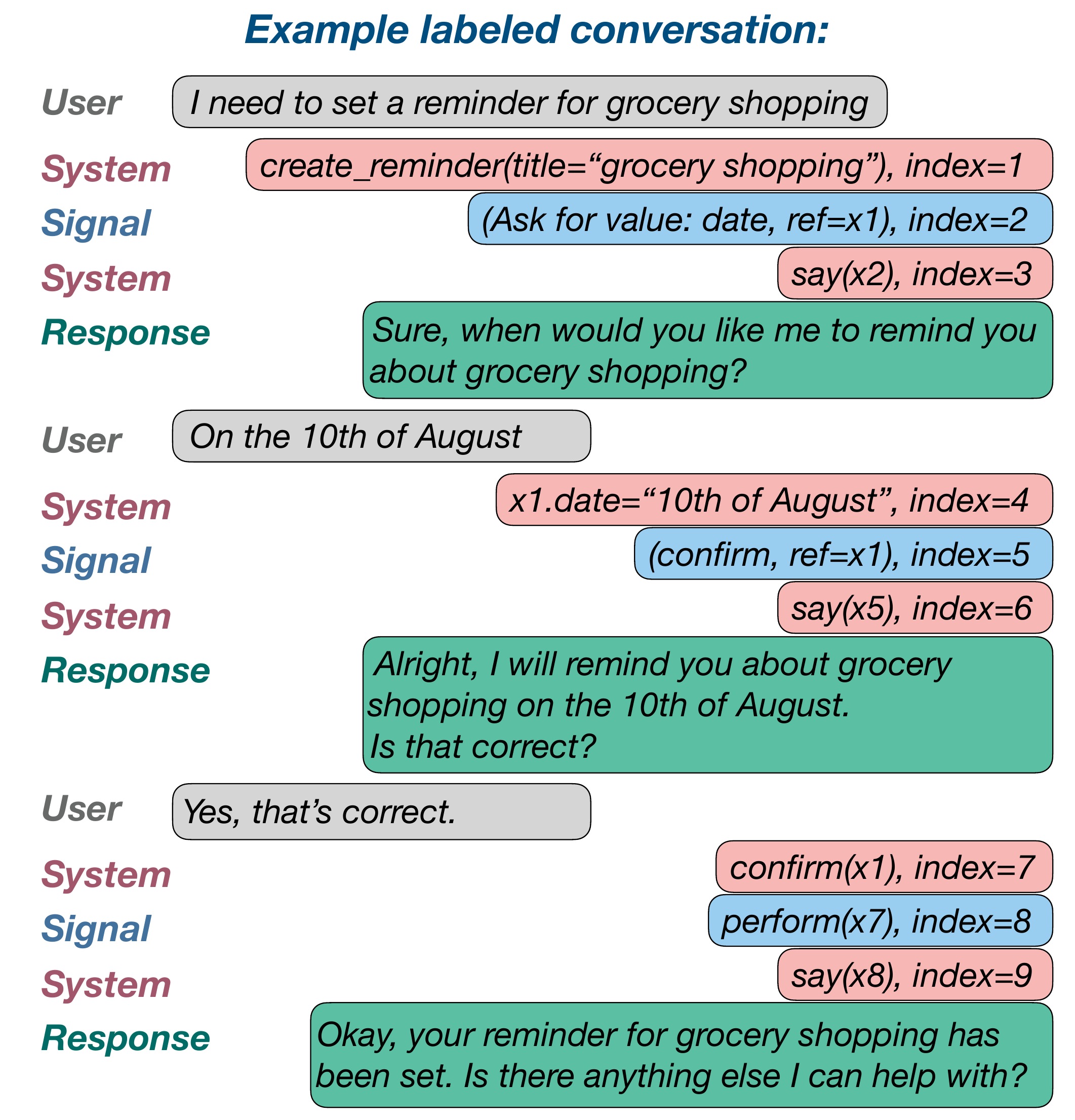}
    \caption{A (simplified) example labelled conversation. Each dialogue contains user, system, signal and response turns.} \label{pirana}

\end{center}
\end{figure}
\subsection{Annotation Scheme and our Mock Back-end}
We apply a concise labelling system to track the states for each intent in a conversation. This labelling system follows a Pythonic syntax, with function calls used to initialise intents and entities when these are first mentioned, and attribute assignments used for any subsequent slot filling operations (see \cref{pirana}).  This schema-based, concise form of semantic labelling is highly convenient, and avoids the need for state tracking for each individual turn.

Our labelling involves four types of turns: 1) User turns; 2) System turns, labelling user intentions; 3) Signal turns, returned after our mock back-end processes the system command; and 4) Response turns, which are natural language responses back to the user. Further details on our labelling schema is provided in \cref{sec:annotation_schema}.

%System turns always follow user turns. In most cases, the first system turn is followed by a signal turn, except when the system decides to immediately call a response with `say()', for example if the user response is irrelevant. Signal turns are then followed by a system turn, which triggers the natural language response turn. The system turns therefore decide when to pass information to the mock back-end, and when to trigger the natural language response. We use the system turns as the targets in this dataset.

%LUCID automatically creates the mock back-end for each intent using the schema generated in steps 1 and 2. This involves generating a Python class to represent the intent in question, which is then instantiated as an object and interacts with the system commands to indicate when mandatory slots have not yet been provided, or when confirmation is still required before the intent can be performed.

\section{Analysis}
\subsection{Diversity of Slots and Intents}
The generated LUCID data contains more intents and slots than existing task-oriented dialogue datasets (\cref{intent_unhappy_path_summary}). Specifically, the dataset contains 100 intents, across 13 domains, with 501 different slots. While the SGD dataset contains more domains than LUCID, these domains are narrower in scope. For example, SGD includes separate domains for buses, taxis, flights and trains, while LUCID has a single transportation domain incorporating intents for each of these areas. The larger number of slots and intents in LUCID illustrates our ability to create diverse and challenging data using LUCID, despite generating a smaller dataset compared to SGD and MultiWOZ (see \cref{reported_size}).

%Considering the size of our dataset, compared to SGD and MultiWoz there are almost twice as many unique slot values, demonstrating the level of variety in our dataset (see \textit{values per turn} in \cref{reported_size}).

%We compare the number of slots within LUCID to PRESTO, SGD and MultiWoZ in \cref{intent_unhappy_path_summary}, showing that after excluding duplicated slots in both PRESTO and SGD, the LUCID generated data has almost twice the number of slots. \cref{sec:remove_dupliates} outlines our approach for removing these duplicated slots, in the case of SGD when the same slots are included in different services, or in PRESTO when slots are counted multiple times if they have slightly different parse trees.

\begin{table*}
\begin{center}
%\resizebox{\textwidth}{!}{
\begin{tabular}{rccccc}
\toprule 
&
 \textbf{\# Domains} & \textbf{\# Intents} & \textbf{Ints per Dom} & \textbf{\# Slots} & 
 \textbf{\# Labelled Unhappy Paths.} 
 \\
\midrule
 PRESTO & - & 34 & - & 303$\dagger$ & 6 \\
 PRESTO-no dup & - & 34 & - & 276$\dagger$ & 6 \\
 SGD & \textbf{20} & 88 & 4.4 & 365 & 0 \\
 SGD-no dup & \textbf{20} & 46 & 2.3 & 240 & 0 \\
 MultiWOZ & 7 & 11 & 1.6 & 35 & 0 \\ % tracked slots
 \midrule
 LUCID & 13 & \textbf{100} & \textbf{7.7} & \textbf{501} & \textbf{9} \\
\bottomrule
\end{tabular}
\end{center}
\caption{Summary statistics of our dataset, displaying the number of domains and intents present, the number of intents per domain, the number of slots present, and the number of explicitly labelled conversational phenomena (unhappy paths). For PRESTO, we consider the 303 slots in English intents ($\dagger$). Unlike \cref{reported_size}, this table considers all splits in the dataset. \cref{sec:remove_dupliates} describes how duplicate slots and intents are removed for SGD and PRESTO.}
\label{intent_unhappy_path_summary}
\end{table*}

As the LUCID dataset was generated primarily to showcase the capabilities of the LUCID data generation system, others are free to use the LUCID system to generate much larger and even more complex datasets. This extensibility is what most clearly distinguishes LUCID from these other data generation efforts.

\subsection{Conversational Phenomena}

LUCID contains a greater number of labelled conversational phenomena than existing dialogue datasets (\cref{intent_unhappy_path_summary}). The recently released PRESTO dataset also contains turn-level annotated phenomena, labelling six types of unhappy paths \citep{PRESTO}. These unhappy paths include in-turn corrections, correcting actions, correcting slot values, code-mixing, disfluencies and cancellations. While half of these phenomena relate to corrections, this is the case for only two of our labelled phenomena, correcting slot values either in-turn or across turns. Instead, we focus on distinguishing between relevant, sensible user replies from cases where a virtual assistant should ask for clarification (rather than using the initial response to populate slot values).

%While not explicitly stated, some specific conversational phenomena can also be derived from existing datasets, such as co-references \cite{dst_trees}, or user corrections \cite{SGD}.

\subsection{Qualitative and Quantitative Analysis} \label{sec:post_processing}
We perform a qualitative analysis on the generated dataset (conducted by one of the paper authors) to thoroughly investigate the dataset quality and identify any issues. This included a manual review of 200 conversations in our dev set, which only highlighted two labelling errors (impacting only 1\% of conversations). In comparison, \citet{eric-etal-2020-multiwoz} identify annotation errors in 40\% of turns in MultiWoz 2.0, demonstrating the relative quality of the LUCID system labels. 

The two labelling errors identified in this review involved: 1) A user mentioning there will be no spoilers in a review, where LUCID correctly assigns the spoiler alert slot value as False, but additionally includes the text `no spoilers in my review' as part of the review itself. 2) LUCID not recognising an in-turn correction by the user, mistakenly including all of the user text (including the correction itself) as part of a slot value. See \cref{sec:quantitative_analysis} for further details. We additionally share a qualitative analysis of our data in \cref{sec:qualiltative_analysis}, highlighting specific areas where our method could be further improved. We include this analysis to further raise the bar for future LLM data generation efforts.

\section{Baseline Results}
\begin{table}
\begin{center}
%\resizebox{\textwidth}{!}{
\begin{tabular}{rccccccc}
\toprule 
 &
 \textbf{Intent acc.} & \textbf{JGA}  \\
\midrule
Test (seen): \\
\midrule
T5-Small & 94.7 & 57.1 \\
T5-Base & 97.9 & 67.5  \\
T5-Large & \textbf{98.7} & \textbf{69.0} \\
\midrule
 Test-OOD (unseen): \\
 \midrule
T5-Small & 95.3 & 22.0 \\
T5-Base & 97.6 & 42.2 \\
T5-Large & \textbf{98.8} & \textbf{45.7} \\
\bottomrule
\end{tabular}
\end{center}
\caption{Results of our baseline model (with retrieval). Full results and evaluation metric descriptions are provided in \cref{sec:baseline_results}.}
\end{table}

We train T5 \cite{T5} and Flan-T5 \cite{FlanT5} baseline models on LUCID, evaluating on both our in-distribution and out-of-distribution test sets. When retrieving intent schemas, a Sentence-BERT \cite{reimers-2019-sentence-bert} model is used to encode the tool name and the last user utterance, choosing the tool with the highest cosine similarity.

As expected, the joint goal accuracy is considerably higher when evaluating on the seen test set compared to the unseen test set, with accuracy scores of 67.5\% and 42.2\% respectively for a T5-base model (see \cref{full_t5_results}). We also isolate the impact of the retrieval model, comparing three scenarios: 1) using our tool retrieval, 2) using an oracle tool retrieval, and 3) including no tools in the prompt. We find that the tool retrieval is not a major weakness of our baseline model (see \cref{tool_retrieval}). %Finally, we experiment with training a Flan-T5-Base model with smaller amounts of training data, finding little benefit from a larger training corpus.
Finally, we evaluate our Flan-T5-base model on each different conversational phenomena (see \cref{sec:conv_phenomena}), highlighting sarcasm, ASR-early end examples, and answering about a different slot as the most challenging phenomena. %On the other hand, cancellations, users delaying giving confirmation, and overheard answers are more easily predicted, and our baseline model makes no mistakes in these cases. 
Full experimentation details and results can be found in \cref{sec:baseline_results}.

\section{Conclusion}

We introduce LUCID, a pipeline of LLM calls which is designed to create high quality and linguistically sophisticated dialogue data. LUCID involves an extensive validation process, including three validator LLMs that discard conversations where there is any disagreement. To demonstrate the quality of the data produced, we generate a seed dataset of 4,277 dialogues, consisting of 92,699 turns, with a wide variety of challenging conversational phenomena. The generated system labels in LUCID prove to be highly accurate, with only 1\% of conversations containing a labelling error. We make our code available to facilitate larger scale, high quality data generation.

\section*{Limitations}

The main limitation of our approach is the cost of using a closed-source LLM. This prevented us from generating a larger number of dialogues or performing more ablation studies to isolate the improvements from specific stages. This cost was driven by our \textit{if in doubt, discard} approach to validation, which prioritised the accuracy and quality of the data produced, at the expense of the computational time and cost involved. While there are also substantial costs associated with high quality manual annotation, in this work we aim to show that an LLM-driven approach to generating high quality data is possible and feasible. We also aim to produce a seed dataset of the highest quality which can be used by practitioners on an on-going basis.

\bibliography{anthology, custom}
\clearpage
\appendix

\begin{figure*}
\begin{center}
    
\includegraphics[width=430pt]{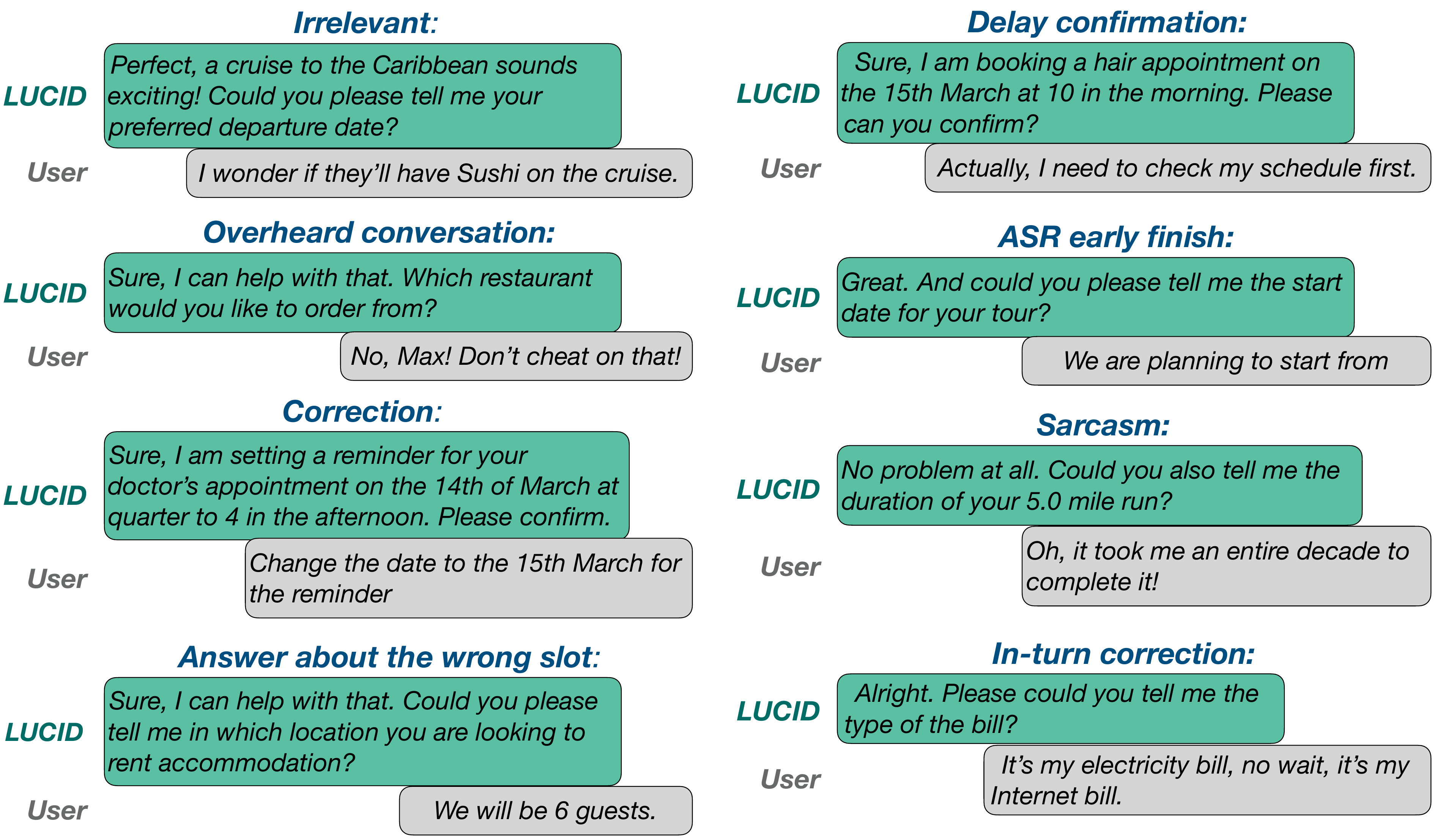} 
    \caption{Examples for eight of the nine challenging conversational phenomena included in the LUCID dataset. We also included `cancellation' examples which are similar to `delay confirmation', resulting in the system not confirming a given intent.} \label{unhappy_paths}

\end{center}
\end{figure*}

\begin{figure}
\begin{center}
\includegraphics[width=\columnwidth]{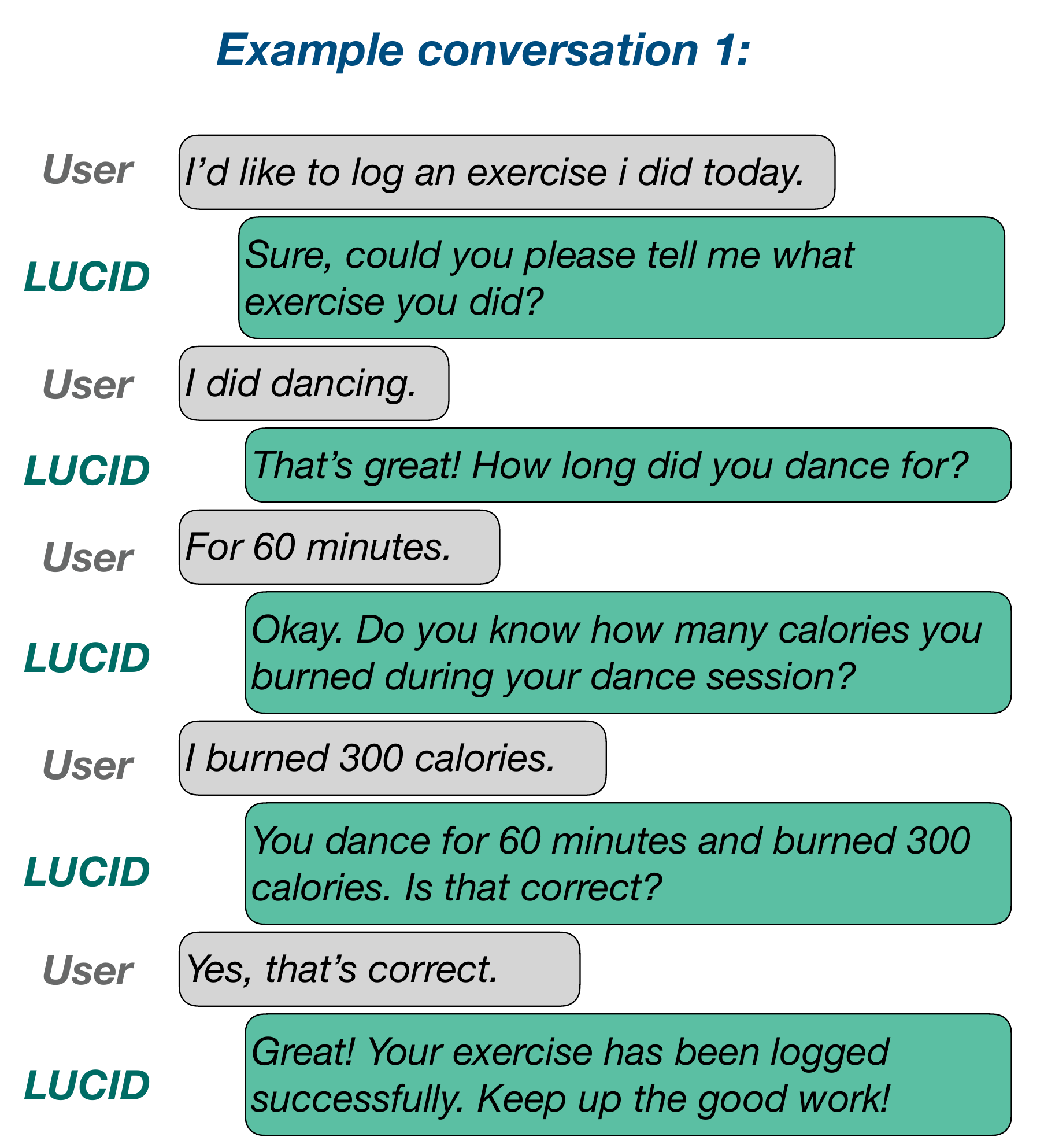} 
    \caption{An example conversation from LUCID (Example \#1). As described in \cref{sec:examples}, we show the first three LUCID conversations to provide an unbiased sample of our generated data.} \label{ex1}
\end{center}
\end{figure}
\section{Related work - Data Generation with LLMs} \label{sec:related_work_llm_gen_data}

\citet{wu2023autogen_llm_to_llm} recently introduce a framework allowing the interaction of multiple, different LLMs, based on the idea that LLMs can solve highly challenging tasks if these tasks are broken into smaller steps. While \citet{wu2023autogen_llm_to_llm} are successful in generating dialogues for a group chat scenario, this does not require the intent and slot labelling needed for task-oriented dialogue. For other NLP tasks, to avoid labelling errors or poor quality data, generating data with LLMs can involve human annotators reviewing the generated utterances \cite{liu2022wanli, wiegreffe-etal-2022-reframing}, or using the generated data as unlabelled data to be used with knowledge distillation \cite{meng2022generating, DBLP:journals/tacl/HeNKH022, stacey2023improving}. Labelled data generation has been successful for other tasks without human input \cite{almost_no_human_input, zero_gen, schick-schutze-2021-generating, west-etal-2022-symbolic, Yuxiang_UCL_paper}, however noise may be an issue for a large proportion of the data \cite{almost_no_human_input, schick-schutze-2021-generating}. 

\section{Conversation Planner Details} \label{sec:planner_sampling} 
To ensure variety in the generated conversations, the planner makes extensive use of sampling, choosing how many intents should be provided in the conversation, which optional slots should be discussed, the conversational phenomena (both happy and unhappy paths) that should be included, and which slots and intents any conversational phenomena should be applied to.

Sampling is also involved in choosing the path of intents that are included in the conversation. The planner is provided with a sample of intents across all domains, before an LLM generates a plausible sequence of intents from this sample (\textit{stage 3}).

We also provide more detail below to describe exactly how the conversation planner creates slot values for the conversation, involving a range of different stages: 

The slot values for the first intent are initially sampled based on the plausible slot values generated for each intent (see \textit{stage 2}), preventing LLMs generating repetitive conversations. An LLM is then asked to make the slots for each intent more realistic and coherent (\textit{stage 4}). This prevents contradictory slot values, for example when a hotel check-out date is before the check-in date. This stage also prevents highly unlikely slot values being over represented, while introducing further variety into the slot values being provided.

For all other intents in the conversation (after the first intent), an LLM generates a plausible reason why the user would want to complete this intent given what has already occurred in the conversation (\textit{stage 5}). Likely slot values are then generated (\textit{stage 6}) based on this context. \textit{Stage 4} is then repeated, increasingly the likelihood that the slot values selected are realistic and coherent for the intent. An LLM is then asked to update the slot values across every intent in the conversation so that these intents are related and consistent (\textit{stage 7}), encouraging more natural conversations that align more closely with human to virtual assistant interactions.

\section{Quantitative Analysis Findings} \label{sec:quantitative_analysis}
\cref{quantitative_analysis} provides the full results from our quantitative analysis of 200 dev examples. This analysis identified two labelling issues within the 200 conversations that were reviewed (conducted by one of the paper authors). 

In the first instance, the user says ``I want to review the film the godfather. I give it a 9 out of 10 and my review is an absolute classic! Great performances and storytelling.. No spoilers in my review.'' The system interprets this as giving the film name (the godfather), the rating (9), the spoiler alert (False), and the review text. However, the system also predicts the text `No spoilers in my review' as part of the review text, when this may not be the case.

In the second instance, the user performs an in-turn correction, changing the value of the additional notes slot for a new hair appointment intent. However, the system predicts this correction as part of the note itself, giving the slot value: `I need a quick haircut, actually make that I'm getting ready for a family reunion photoshoot and want a new haircut.'

\cref{quantitative_analysis} also measures some of the issues reported in the qualitative analysis, including: 1) how many times the natural language generated response (NLG) indicates an intent was performed before confirmation was given, 2) when the NLG does not follow the system predictions, 3) unrealistic slot values, 4) unrealistic combination of slots mentioned for an intent, 5) when the first choice string span was not selected by LUCID, and 6) when the user or NLG does not fully understand the purpose of the intent. While previous work does not report similar metrics, we publish these figures with the aim of raising the bar for future data generation efforts.

In the case of point 5) considering when the first choice string span was not selected, we find that only 8.5\% of conversations have a string slot value that does not match the reviewer's first choice. However, these differences are subjective and subtle, with over half of the cases concerning whether to include `the' before a date (e.g. `the 5th of March' vs `5th of March').

\begin{table*}
\begin{center}
%\resizebox{\textwidth}{!}
\begin{tabular}{lcc}
\toprule 
 \textbf{Issue name} & \textbf{Prevalence} \\
\midrule
 NLG indicates intent performed before confirmation & 4.5\% \\
 NLG does not follow system prediction & 1\% \\
 Unrealistic slot values & 6\% \\
 Unrealistic combination of slots mentioned for an intent & 3.5\% \\ % tracked slots
 First choice string span not selected & 8.5\% \\
 User or NLG does not fully understand purpose of the intent & 6.5\% \\

\bottomrule
\end{tabular}
\end{center}
\caption{Quantitative analysis from 200 dialogues in our development set. We report these six metrics in addition to the system label accuracy figure of 1\% provided in \cref{sec:post_processing}.}
\label{quantitative_analysis}
\end{table*}

\section{Qualitative Analysis} \label{sec:qualiltative_analysis}

We perform a qualitative analysis on our dev set, understanding potential limitations of our data, and suggesting ways these could be mitigated for future generation efforts.

\textbf{Finding 1)} \textbf{The natural language responses from the model do not always reflect the system labels that have been predicted.} We observed that a correct system label can be accompanied by a natural language response that does not reflect the correct system prediction. We noticed this for sarcastic responses, where only the system label and not the natural language response reflected the user's sarcasm. We choose to manually review the turns labelled as sarcastic, filtering out 4 dialogues. However, our quantitative analysis on the dev set highlights this as an issue beyond sarcastic turns, with natural language responses not faithfully following the system label predictions in 1\% of conversations.

As a related issue, the natural language response can also suggest that an intent has been performed before confirmation is given by the user. Informed by this finding, we filter out conversations when an intent was not performed, unless there was a cancellation signal provided by the user (removing 79 conversations). After this filtering, our quantitative analysis finds that 4.5\% of conversations include responses that suggest an intent has been performed before confirmation is given. However, in each case there was no impact on the conversation beyond the phrasing of the natural language response. As LUCID prioritises validating the system labels, we do not implement validation checks on the natural language response. Introducing additional validation for the natural language responses is likely to also improve their quality.

\textbf{Finding 2)} \textbf{The planner's choice of slots and their corresponding values can sometimes be unrealistic}. While a strength of LUCID is the realistic and varied slot values used in conversations, this is not always the case. We also notice that the choice of slots included in a conversation is not always realistic. For example, you would not usually give the start time, end time in addition to specifying the duration of a swimming lesson. Our quantitative analysis identifies that 6\% of conversations contain at least one unrealistic slot value, while 3.5\% of conversations include an unrealistic combination of slots. The unrealistic slot combinations demonstrate a limitation to our sampling approach, where we randomly sample which optional slots should be included in each conversation. This issue could be overcome with an additional LLM stage responsible for deciding if the slot combination provided is realistic or not.

\textbf{Finding 3)} \textbf{The user does not always understand the purpose of the intent}. For example the user may ask `can you find my favorites from yesterday?', when it is not clear if the user understands what a `favorite' is. This is a consequence of our conversation plans telling users which intent should be performed, without also providing a description. The quantitative analysis finds that in 6.5\% of conversations, either the user or the natural language response does not fully understand an intent, suggesting that descriptions should be included for future data generation work.

\textbf{Finding 4)} \textbf{The system command labelling is consistently high quality, with few labelling mistakes}. We quantify this finding with our quantitative analysis of 200 conversations in the development set, which finds only 1\% of examples when the system label is not correct. More detail on the two system labelling errors identified are provided in
\cref{sec:quantitative_analysis}.

%6\% of slot values were considered unrealistic, for example running 3.2 miles in 2 hours, or specifying the time that a program is shown on Netflix. Additionally, 3.5\% of conversations did not contain a realistic combination of slots. 

%We find that in 6.5\% of conversations, either the user or the natural language response does not fully understand the intent. For example the user may ask for `Can you find my favourites from yesterday?', when it is not clear they understand what a `favorite' is. This is a consequence of only providing users with the intent command that should be performed, without also providing its description. 
%Finally, 1\% of the natural language responses did not reflect the system commands. In one case, the response says that a reminder has been set for a note when it wasn't. In another case, the response uses a slightly different slot value, referring to 'Coach John Smith' rather than the predicted slot value of 'John Smith.'

% At some point, should say which datasets follow an explicitly defined schema for intents.

\section{Validation and Post-Processing} \label{sec:validation_info}

We introduce additional validation, ensuring turns that include our challenging conversational phenomena are correctly predicted by our LLMs. When the conversation rules instruct a user to introduce a specific conversational phenomenon for a certain slot value, the user is instructed to also provide a signal (in the form of a special token) to show that this unhappy path is being performed. We then use this signal for validation purposes, ensuring that the following system command matches the expected response for this phenomenon. However, we do not provide these special tokens to the system which interprets and labels the user request; these would not be available to a real virtual assistant and we find that including them during data generation can result in unrealistic target labels (e.g. if a user's `irrelevant' response accidentally constitutes a plausible slot value).

A range of post-processing rules are also introduced after our qualitative analysis. We filter conversations where a slot is corrected during the conversation, but where there is no correctional signal provided by the user (as described above, a signal is provided by the user for each complex conversational phenomena which is used purely for validation purposes). This filtering process removes instances where a slot was first mentioned by the user without giving a value, with the system incorrectly assigning a slot value from this turn (filtering 123 conversations). We perform additional filtering to remove empty string slot values (removing 27 conversations), and any instances where the system turn predicts a hint, as hints should only occur in Signal turns. There were 172 occurrences when a hint was predicted by the system, although in almost all cases these conversations were already filtered by another post-processing filter.

In total, 56\% of conversations pass all of our validation checks. To avoid wasting valuable conversational data, we salvage the prefix of an aborted conversation up to the point where the validation error was identified\footnote{To avoid overly short conversations, we do this only if at least one intent has been performed already or at least 10 turns have occurred}. In these cases, we truncate the conversation, sampling LLM generated natural language responses that justify interrupting the conversation.

\section{Annotation Schema Detail} \label{sec:annotation_schema}

An important part of our annotation schema is the order of the turns, and how system turns trigger the natural language responses. This section provides more detail on these points.

System turns always follow user turns. In most cases, the first system turn is followed by a signal turn, except when the system decides to immediately call a response with `say()', for example if the user response is irrelevant. Signal turns are then followed by a system turn, which triggers the natural language response turn. The system turns therefore decide when to pass information to the mock back-end, and when to trigger the natural language response. We use the system turns as the targets in this dataset.

LUCID automatically creates the mock back-end for each intent using the schema generated in steps 1 and 2. This involves generating a Python class to represent the intent in question, which is then instantiated as an object and interacts with the system commands to indicate when mandatory slots have not yet been provided, or when confirmation is still required before the intent can be performed. The outputs of the mock back-end are represented by the signal turns described above. 
%The system turns therefore interact with the mock back-end, which in turn produce a signal turn, with a subsequent signal turn deciding what the resulting action should be (e.g. what the natural language response should say).

\section{Baseline Results} \label{sec:baseline_results} 
We train six different baseline models on the LUCID training data (T5-small, T5-base, T5-large, Flan-T5-small, Flan-T5-base and Flan-T5-large models). Each model is evaluated on the test set for seen intents and the OOD test set for unseen intents (see \cref{full_t5_results}). We additionally experiment with training our Flan-T5-base baseline on varying amounts of training data (see \cref{training_data_size}). Details about about the choice of hyper-parameters can be found in \cref{sec:training_setup}.

To measure performance on our generated LUCID data, we consider five performance metrics: joint goal accuracy, intent accuracy, fuzzy slot accuracy, exact match accuracy (between user turns), and exact match accuracy for an entire dialogue. 

For joint goal accuracy, we consider a fuzzy matching score for string slot values. As many system turns involve a \textit{say} command, a joint goal accuracy figure is only calculated for turns where a value is predicted (or contained in the system labels). Additionally, we consider the goal state of all intents in the conversation, rather than considering different states for different intents or domains separately.

We also use fuzzy matching for our slot accuracy measure, which is a joint accuracy measure across all the slot values provided in a single system turn (when any slot values are predicted, or when they are included in the labels). We additionally introduce exact match metrics that consider the accuracy of all system commands, not just those that refer to intent and slot values (for example, including `say' commands). We introduce two exact match metrics - \textit{exact match (turn)} considers whether all predicted system commands between two user turns exactly match their labels, while \textit{exact match (conversation)} considers whether every predicted system command in a conversation matches with the system labels.

The exact match between user turns is measured for a Flan-T5-base model for each conversational phenomena, both in the seen and unseen (OOD) test sets (see \cref{unhappy_path_numbers}). We use this metric because some conversational phenomena involve an assignment, which is then followed by a `say' command. As predicting a `say' command following an assignment is not challenging for the model, we find that using the exact match between user turns metric provides the fairest comparison. The most challenging phenomena for our Flan-T5-base model are ASR-early end, sarcasm and answering about another slot phenomena (see \cref{unhappy_path_numbers}), although predictions for ASR-early end are substantially worse for the seen intents. A number of phenomena appear to be less challenging than examples with no unhappy paths (see `None' in \cref{unhappy_path_numbers}), particularly for the OOD test set. This occurs because many phenomena do not involve any slot assignment, which becomes more challenging in the OOD test set.

For the tool retrieval, as gold system labels for previous turns are seen in the prompt conversation history, we retrieve all tools that have been mentioned in an oracle history up to that point.

Our baseline models are fine-tuned using the following prompt: 
 ``You are a smart AI assistant who is responsible for writing system commands to describe what the user has asked for. Your job is to write the next system command based on the latest user turn, considering the conversation so far.'' When using tool retrieval, the following text is added ``Information about the following tools may help:'', before providing the retrieved intent alongside intents from the conversation history. Finally, a single in-context example is provided to the model (see our code for further details). 

\begin{table*}
\begin{center}
%\resizebox{\textwidth}{!}{
\begin{tabular}{rccccccc}
\toprule \textbf{Conv. phenomena}
& \textbf{Total} &
 \textbf{Train} &
 \textbf{Dev} &
  \textbf{Test} & \textbf{Test-OOD} & 
 \textbf{Test} & \textbf{Test-OOD} \\
 & \# & \# & \# & \# & \# & Acc. & Acc.
\\
\midrule
 Cancellation & 12 & 5 & 2 & 3 & 2 & 100 & 100 \\
 ASR-early end & 58 & 41 & 7 & 7 & 3 & 43 & 100 \\
 Sarcasm & 63 & 46 & 3 & 8 & 6 & 75 & 67 \\
 Delay confirmation & 76 & 53 & 7 & 4 & 12 & 100 & 100 \\
 Answer about another slot & 113 & 75 & 13 & 11 & 14 & 64 & 43 \\
 Irrelevant answer & 163 & 116 & 18 & 15 & 14 & 93 & 93 \\
 Overheard answer & 203 & 153 & 17 & 23 & 10 & 100 & 100 \\
 In-turn correction & 215 & 145 & 27 & 25 & 18 & 80 & 72 \\
 Correction & 250 & 166 & 28 & 31 & 25 & 90 & 81 \\
 None & 3,200 & 2,279 & 307 & 252 & 362 & 82 & 56 \\
 \midrule
 Conv. w/ 1+ unhappy path & 1,077 & 754 & 108 & 119 & 96 & - & - \\
 Total conversations & 4,277 & 3,033 & 415 & 371 & 458 & - & - \\
 \midrule
 \% Conversations unhappy & 25.2\% & 24.9\% & 26.0\% & 32.1\% & 21.0\% & - & - \\
\bottomrule
\end{tabular}
\end{center}
\caption{Total number of each conversational phenomenon within each split of our dataset. While there are few conversations for cancellation, this behaves similarly to the `delay confirmation' phenomenon. We also show the exact match (Turn) metric for each conversational phenomena from a T5-Flan-base baseline model (details of the metric are provided in section \cref{sec:baseline_results}).}
\label{unhappy_path_numbers}
\end{table*}

\section{Intent Generation} \label{sec:intent_generation} In total, 54 intent descriptions were provided, with a single intent removed for data quality reasons. The removed intent involved the user asking the virtual assistant to start watching a television channel, giving specific start and end times for when they want to start watching the channel. As this is an unrealistic scenario (a user would want to start watching a television channel straight away), the intent was removed. 

A transactional intent was created for each of the remaining 53 descriptions. LUCID then creates a query intent corresponding to each transactional intent. The query intent returns entities that would be created using the corresponding transactional intent.

Some query intents were merged together (this happens when the corresponding transactional intents had the same entity names - one of the intent properties generated by the LLM). As a result, there are 6 fewer query intents than transactional intents, resulting in a total of 100 intents. Each of the following pairs of transactional intents returned the same entity names, and so their corresponding query intents were combined: \textit{add\_tv\_program\_to\_favorites} and \textit{add\_artist\_to\_favorites} (both of which return `favorites' entities), \textit{set\_timer} and \textit{set\_alarm} (both of which return `alarm' entities), 
\textit{book\_nails\_appointment} and \textit{book\_spa\_appointment} (both of which return `appointments'), 
\textit{order\_supermarket\_shop} and \textit{order\_takeaway} (both of which return `orders'),
\textit{add\_song\_to\_favorites} and \textit{play\_song} (both of which return `songs'), and \textit{review\_film} and \textit{review\_restaurant} (both of which return `reviews').

Each of the 100 intents used for our data generation are listed in \cref{list_all_intents} and \cref{list_all_intents_31}. These tables list each transactional intent, along with its corresponding query intent. We also provide the human authored descriptions for each intent that were initially provided to LUCID.

\section{Slot Duplication within PRESTO and SGD} \label{sec:remove_dupliates}

SGD report 214 slots in their training data \cite{SGD}, corresponding to 365 slots across all dataset splits (see \cref{intent_unhappy_path_summary}). This counts slots with exactly the same names in the same domains within different services, which we consider to be duplicated slots (although the allowed slot values may change in each case).
As a result, we provide a more direct comparison to SGD without this slot and intent duplication across services (see `SGD-no dup' in \cref{intent_unhappy_path_summary}).

For PRESTO, we consider the 303 slots present in the English split of the dataset \citep{PRESTO}. However, as the semantic annotations in PRESTO are represented in parse-trees, slots are counted multiple times if their paths are different. We find the number of English slots in PRESTO reduces to 276 without this duplication.

When considering the total number of slots in PRESTO, the same slot can be counted multiple times depending on its position in the labelled parse trees. For example, the \textit{Send\_digital\_object} intent includes \textit{bcc} and \textit{cc} slots. Both of these slots can be a \textit{Personal\_contact} entity, which contains a \textit{person} slot. In this case, the \textit{person} slot within \textit{Personal\_contact} would be counted twice if it was contained within either the \textit{bcc} or \textit{cc} slots. Removing this slot duplication reduces the number of English slots in PRESTO from 303 to 276.

Note, we consider the v.2.2 of MultiWOZ for our comparison, as this version explicitly states the intents present in the dataset. 

\section{Example Dialogues} \label{sec:examples}
\begin{figure*}
\begin{center}
\includegraphics[width=420pt]{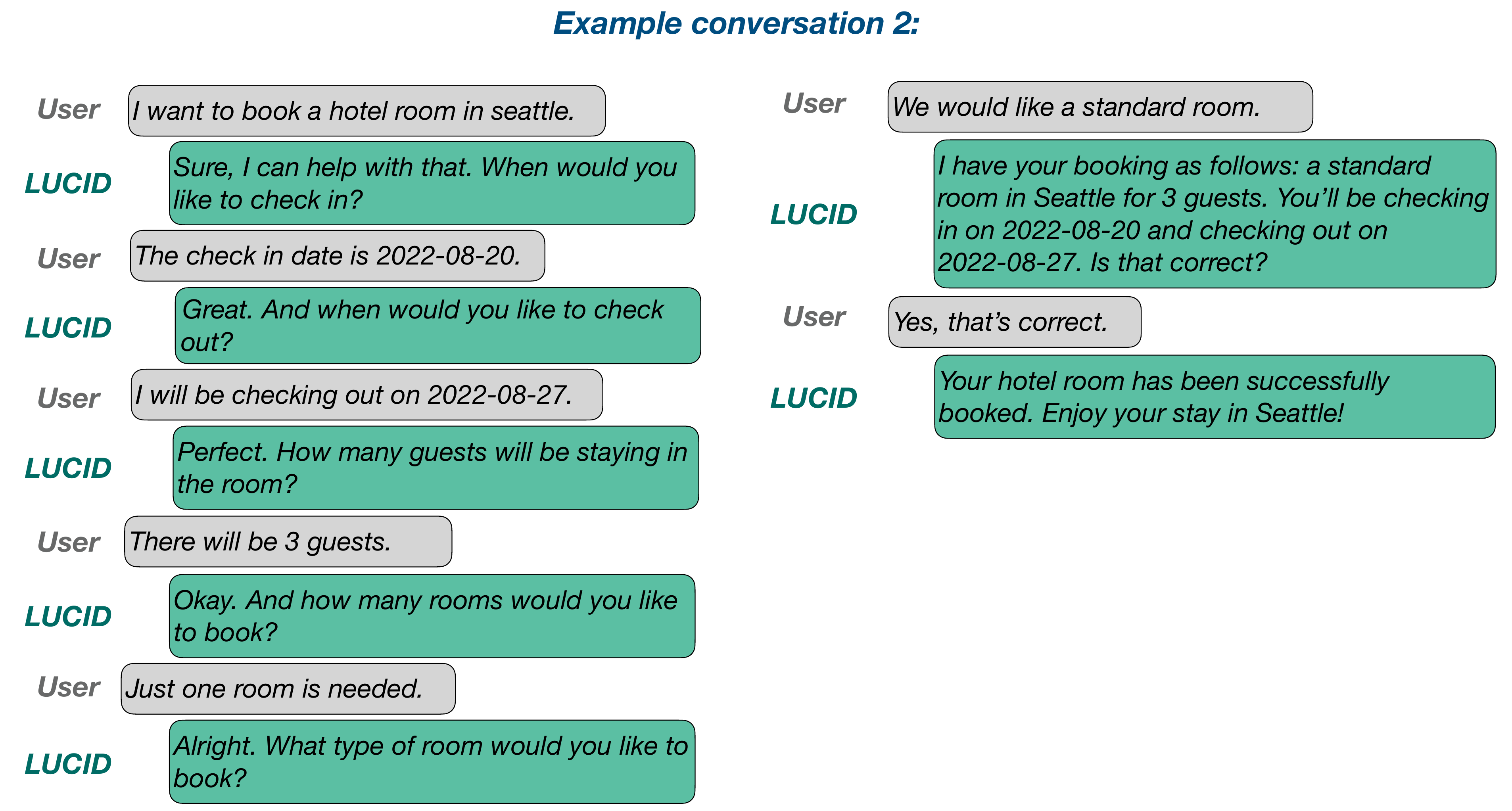} 
    \caption{An example conversation from LUCID (Example \#2). As described in \cref{sec:examples}, we show the first three examples to provide an unbiased sample from our generated data.} \label{ex2}
\end{center}
\end{figure*}

\begin{figure*}
\begin{center}
\includegraphics[width=420pt]{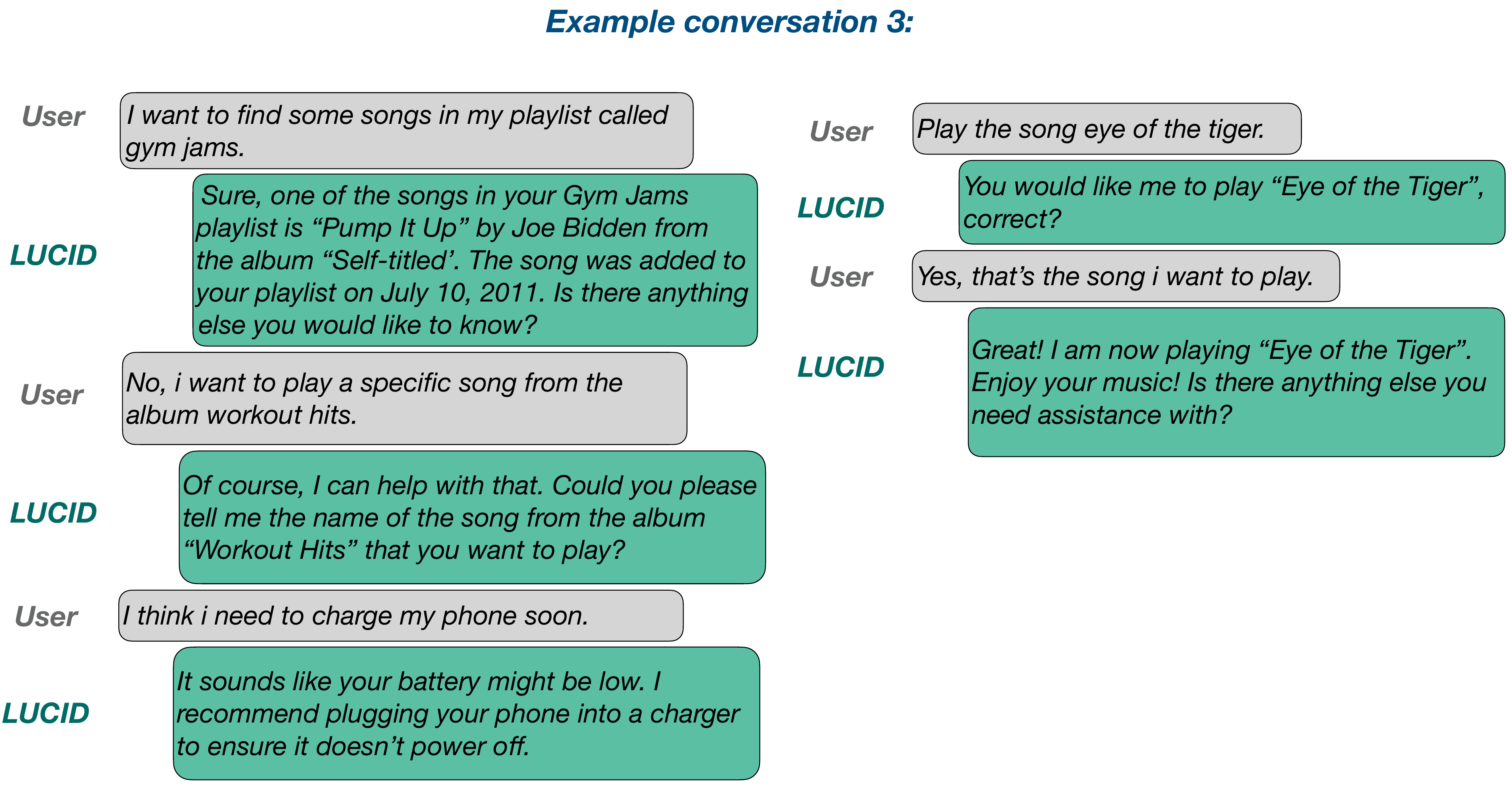} 
    \caption{An example conversation from LUCID (Example \#3). As described in \cref{sec:examples}, we show the first three examples to provide an unbiased sample from our generated data.} \label{ex3}
\end{center}
\end{figure*}

In addition to the examples provided in \cref{single_extract} and \cref{pirana}, we provide three additional examples of the LUCID generated conversations. To provide an unbiased sample of our conversations, we show the first three dialogues in the dataset (see \cref{ex1}, \cref{ex2} and \cref{ex3}). We also show examples of each of the unhappy paths used in our dataset (see \cref{unhappy_paths}).

\section{Modelling Setup, Parameters, Computing Setup} \label{sec:training_setup}
For each baseline experiment, we train for 3 epochs. This was selected as a hyper-parameter based on the development set loss after training for 1-10 epochs for our Flan-T5-base baseline. We use a learning rate of 5$\times10^{-5}$, with a linear learning schedule, a batch size of 2 with 8 gradient accumulation steps. For experiments with reduced training data, we train for more epochs as the training data is increased (with epochs inversely proportional to the size of the training data, allowing for a fair comparison). For each LLM call, we use a temperature value of 0.7.

Our baseline models have the following number of parameters: T5-small and Flan-T5-small (60 million parameters), T5-base and Flan-T5-base (220 million parameters), T5-large and Flan-T5-large (770 million parameters). We train our models with V100 GPUs, with our combined baseline experiments training for approximately 80 GPU hours. 

\begin{table*}
\begin{center}
%\resizebox{\textwidth}{!}{
\begin{tabular}{rccccccc}
\toprule 
 &
 \textbf{DSTC2} & \textbf{WOZ2.0} & \textbf{FRAMES} & \textbf{M2M} & 
 \textbf{MultiWOZ} & \textbf{SGD} & \textbf{LUCID}
 \\
\midrule
 \# domains & 1 & 1 & 3 & 2 & 7 & \textbf{16} & 12 \\
 \# dialogues & 1,612 & 600 & 1,369 & 1,500 & 8,438 & \textbf{16,142} & 3,033 \\
 \# turns & 23,254 & 4,472 & 19,986 & 14,796 & 113,556 & \textbf{329,964} & 65,217 \\
 Turns per dialogue & 14.49 & 7.45 & 14.60 & 9.86 & 13.46 & 20.44 & \textbf{21.50} \\
 %Tokens per turn & 8.54 & 11.24 & 12.60 & 8.24 & 13.13 & 9.75 & x\\
 %Total unique tokens & 986 & 2,142 & 12,043 & 1,008 & 23,689 & 30,352 & x\\
 No. of slots & 8 & 4 & 61 & 13 & 24 & 214 & \textbf{432}\\
 No. of slot values & 212 & 99 & 3,871 & 138 & 4,510 & 14,139 & 4,701 \\
 Values per turn & 0.009 & 0.02 & \textbf{0.2} & 0.009 & 0.04 & 0.04 & 0.07 \\
\bottomrule
\end{tabular}
\end{center}
\caption{Reported statistics for LUCID, and related datasets for task-oriented dialogue. All statistics refer to the training split of the datasets, except for Frames which reports figures for all splits. Compared to previous dialogue datasets, LUCID has considerably more slots, and more turns per dialogue. There are also more possible slot values per turn than either MultiWoZ or SGD. The number of turns in LUCID refer to User, System, Signal and Response turns.}
\label{reported_size}
\end{table*}

\begin{table*}
\begin{center}
%\resizebox{\textwidth}{!}{
\begin{tabular}{rccccccc}
\toprule 
 &
 \textbf{Intent acc.} & \textbf{Joint goal acc.} & \textbf{Slot acc.} & \textbf{Match (turn)} & \textbf{Match (conv.)} \\
\midrule
Test (seen): \\
\midrule
T5-Small & 94.7 & 57.1 & 69.8 & 74.5 & 30.5 \\
T5-Base & 97.9 & 67.5 & 76.6 & 82.1 & 44.2 \\
Flan-T5-Base & 97.9 & 69.7 & 77.6 & 82.6 & 45.8 \\
T5-Large & \textbf{98.7} & 69.0 & 77.9 & 83.2 & 46.9 \\
Flan-T5-Large & 98.5 & \textbf{69.7} & \textbf{78.5} & \textbf{83.5} & \textbf{47.4} \\

\midrule
 Test-OOD (unseen): \\
 \midrule
T5-Small & 95.3 & 22.0 & 38.0 & 46.2 & 6.3 \\
T5-Base & 97.6 & 42.2 & 61.4 & 58.5 & 10.3 \\
Flan-T5-Base & 97.6 & 41.3 & 61.2 & 57.0 & 7.6 \\
T5-Large & \textbf{98.8} & 45.7 & 64.1 & \textbf{60.2} & \textbf{11.4} \\
Flan-T5-Large & 98.6 & \textbf{53.2} & \textbf{66.6} & 59.9 & 10.3 \\
\bottomrule
\end{tabular}
\end{center}
\caption{Results of our baseline model trained for 3 epochs, using a SentenceBERT retrieval model. Each evaluation metric is described in more detail in \cref{sec:baseline_results}. Results are from a single seed in each case.}
\label{full_t5_results}
\end{table*}

\begin{table*}
\begin{center}
%\resizebox{\textwidth}{!}{
\begin{tabular}{rccccccc}
\toprule 
 &
 \textbf{Intent acc.} & \textbf{Joint goal acc.} & \textbf{Slot acc.} & \textbf{Match (turn)} & \textbf{Match (conv.)} \\
\midrule
Test (seen): \\
\midrule
No tools & 97.4 &  66.1 & 75.9 & 81.6 & 43.7 \\
w/ retrieval & 97.9 & \textbf{69.7} & 77.6 & 82.6 & \textbf{45.8} \\
Oracle tools &\textbf{99.1}& 69.3 & \textbf{78.1} & \textbf{83.1} & \textbf{45.8} \\

\midrule
 Test-OOD (unseen): \\
 \midrule
No tools & 87.1 & 32.8 & 55.7 & 53.2 & \textbf{8.3} \\
w/ retrieval & 97.6 & \textbf{41.3} & 61.2 & 57.0 & 7.6 \\
Oracle tools & \textbf{99.4} & 40.8 & \textbf{61.3} & \textbf{57.4} & 6.6 \\
\bottomrule
\end{tabular}
\end{center}
\caption{Results of a T5-Flan-base model with our tool retrieval, using oracle tools, and with no tools provided in the prompt. Each evaluation metric is described in more detail in \cref{sec:baseline_results}. Results are from a single seed in each case.}
\label{tool_retrieval}
\end{table*}

\begin{table*}
\begin{center}
%\resizebox{\textwidth}{!}{
\begin{tabular}{rccccccc}
\toprule 
 \textbf{\# Training ex}. &
 \textbf{Intent acc.} & \textbf{Joint goal acc.} & \textbf{Slot acc.} & \textbf{Match (turn)} & \textbf{Match (conv.)} \\
\midrule
Test (seen): \\
\midrule
125 & 88.8 & 29.3 & 49.1 & 57.1 & 10.2 \\
250 & 91.0 & 37.5 & 57.7 & 65.2 & 15.6 \\
500 & 91.9 & 51.2 & 64.9 & 70.6 & 22.6 \\
1k & 94.3 & 58.9 & 70.5 & 75.2 & 29.1 \\
2k & 96.4 & 62.6 & 73.8 & 78.5 & 37.2 \\
4k & 96.7 & 65.0 & 75.0 & 80.3 & 40.4 \\
8k & 97.3 & 66.8 & 76.2 & 81.1 & 42.3 \\
16k & 97.6 & 66.4 & 76.5 & 81.9 & 43.9 \\
%20k & & & & &  \\
Full (24,786) & 97.9 & 69.7 & 77.6 & 82.6 & 45.8 \\
\midrule
 Test-OOD (unseen): \\
 \midrule
125 & 92.7 & 16.7 & 32.3 & 34.9 & 4.1 \\
250 & 93.5 & 20.5 & 43.9 & 44.6 & 4.8 \\
500 & 95.5 & 26.3 & 49.7 & 48.5 & 6.1 \\
1k & 96.7 & 34.3 & 56.4 & 54.2 & 9.6 \\
2k & 96.9 & 32.0 & 56.5 & 53.9 & 5.9  \\
4k & 97.0 & 37.0 & 59.5 & 57.0 & 6.8 \\
8k & 97.2 & 35.5 & 58.6 & 56.5 & 8.5 \\
16k & 97.4 & 32.7 & 57.2 & 55.9 & 7.0 \\
%20k & & & & & \\
Full (24,786) & 97.6 & 41.3 & 61.2 & 57.0 & 7.6 \\
\bottomrule
\end{tabular}
\end{center}
\caption{Results of a T5-Flan-base model trained with varying amounts of training data (count of the system turns provided). Each evaluation metric is described in more detail in \cref{sec:baseline_results}.}
\label{training_data_size}
\end{table*}

\begin{table*}
\begin{center}
%\resizebox{\textwidth}{!}{
\begin{tabular}{lll}
\toprule
Transactional intents (1-30) & Corresponding query & Intent description \\
\midrule
add\_artist\_to\_favorites & find\_favorites & Add artist to favourites \\
add\_event & find\_events & Add event \\
add\_payment\_card & find\_payment\_cards & Add payment card \\ 
add\_restaurant\_to\_favorites & find\_favorite\_restaurants & Add restaurant to favourites \\
add\_song\_to\_favorites & find\_songs & Add song to favourites \\
add\_to\_favourites & find\_favourite\_pages & Add a page to favourites \\
add\_tv\_program\_to\_favorites & find\_favorites & Add a TV program to favourites \\
add\_user & find\_users & Add user with access to calendar \\
block\_sender & find\_blocked\_senders & Block sender \\
book\_bus\_ticket & find\_bus\_tickets & Book a bus ticket \\
book\_city\_tour & find\_city\_tours & Book a city tour \\
book\_cruise & find\_cruises & Book cruise \\
book\_flight & find\_flights & Book a flight \\
book\_guide & find\_guides & Book a guide \\
book\_hair\_appointment & find\_hair\_appointments & Book hair appointment \\
book\_hotel\_room & find\_hotel\_rooms & Book a hotel room \\
book\_massage & find\_massages & Book a massage \\
book\_nails\_appointment & find\_appointments & Book appointment to do nails \\
book\_pedicure & find\_pedicures & Book a pedicure \\
book\_spa\_appointment & find\_appointments & Book a spa appointment \\
book\_swimming\_lesson & find\_lessons & Book swimming lesson \\
book\_taxi & find\_taxis & Book a taxi \\
book\_train\_journey & find\_train\_journeys & Book a train journey \\
book\_triathlon & find\_triathlons & Book triathlon \\
buy\_film\_tickets & find\_film\_tickets & Buy film tickets \\
create\_direct\_debit & find\_direct\_debits & Create direct debit \\
create\_playlist & find\_playlists & Create playlist \\
create\_reminder & find\_reminders & Create a reminder \\
create\_workout\_regime & find\_workouts & Create workout regime \\
log\_exercise & find\_exercises & Log exercise \\
\bottomrule
\end{tabular}
\end{center}
\caption{Each transactional intent (1-30), alongside its respective query intent, and the description provided to LUCID that was used to generate the intent.}
\label{list_all_intents}
\end{table*}

\begin{table*}
\begin{center}
%\resizebox{\textwidth}{!}{
\begin{tabular}{lll}
\toprule
Transactional intents (31+) & Corresponding query & Intent description \\
\midrule
make\_song\_recommendation & find\_recommendations & Make song recommendation \\
open\_web\_page & find\_web\_pages & Open an internet page in a web browser \\
order\_coffee & find\_coffee\_orders & Order coffee \\
order\_supermarket\_shop & find\_orders & Order supermarket shop \\
order\_takeaway & find\_orders & Order takeaway \\
pay\_bill & find\_bills & Pay bill \\
play\_audiobook & find\_audiobooks & Play audiobook \\
play\_film & find\_films & Play film on streaming service \\
play\_podcast\_episode & find\_podcast\_episodes & Play a podcast episode \\ 
play\_song & find\_songs & Play a song \\
rent\_accommodation & find\_accommodations & Rent accommodation \\
rent\_car & find\_cars & Rent a car \\
reserve\_table & find\_reservations & Reserve a table \\
review\_film & find\_reviews & Review film \\
review\_restaurant & find\_reviews & Review a restaurant \\
send\_email & find\_emails & Send an email \\
send\_invoice & find\_invoices & Send invoice \\
send\_message & find\_messages & Send a message \\
set\_alarm & find\_alarms & Set an alarm \\
set\_timer & find\_alarms & Set a timer \\
set\_volume & find\_volume & Set the volume \\
transfer\_money & find\_transactions & Transfer money \\
write\_note & find\_notes & Write a note \\
\bottomrule
\end{tabular}
\end{center}
\caption{Each transactional intent (31+), alongside its respective query intent, and the description provided to LUCID that was used to generate the intent.}
\label{list_all_intents_31}
\end{table*}

\end{document}